\documentclass[letterpaper]{article} 
\usepackage{aaai25}  
\usepackage{times}  
\usepackage{helvet}  
\usepackage{courier}  
\usepackage[hyphens]{url}  
\usepackage{graphicx} 
\usepackage{booktabs}
\usepackage{amsmath}
\usepackage{amsfonts} 
\usepackage{colortbl}
\usepackage{subfigure}
\usepackage{multirow}
\usepackage{caption}
\usepackage{subcaption}
\usepackage{tabularx}
\usepackage[table]{xcolor}
\usepackage{mathtools}
\usepackage[T1]{fontenc}
\usepackage{natbib}  
\usepackage{caption} 
\usepackage{algorithm}
\usepackage{algorithmic}

\usepackage{newfloat}
\usepackage{listings}
\DeclareCaptionStyle{ruled}{labelfont=normalfont,labelsep=colon,strut=off} 

\title{Calibrated Disambiguation for Partial Multi-label Learning}
\author{
  Zhuoming Li \textsuperscript{\rm 1,2}, 
  Yuheng Jia \textsuperscript{\rm 1,2}\thanks{Corresponding author}, 
  Mi Yu \textsuperscript{\rm 2}, 
  Zicong Miao \textsuperscript{\rm 2}\\
}
\affiliations{
  \textsuperscript{\rm 1} School of Computer Science and Engineering, Southeast University \\
  \textsuperscript{\rm 2} China Telecom Cloud Computing Corporation, Beijing 100088, China \\
  \{leezhuoming, yhjia\}@seu.edu.cn; \{yumi2.cq, miaozc\}@chinatelecom.cn

}

\usepackage{bibentry}

\begin{document}

\maketitle

\begin{abstract}
  Partial multi-label learning (PML) aims to train a classifier on datasets whose instances are over-annotated with not only relevant labels but also irrelevant labels, which is common when datasets are collected from crowd-sourcing platform. Existing works primarily approach it from a curriculum learning perspective, leveraging the memorization effect to disambiguate noisy labels and produce robust predictions. However, these methods are based on non-adaptive weighting functions and lack theoretical guidance for optimal weighting. To overcome these issues, a calibrated disambiguation model named PML-CD is proposed. 
  We first formulate the optimal weighting function for curriculum-based disambiguation, which is equivalent to the calibration of the model's predicted confidences, thus providing guidance for curriculum designing. To obtain the optimal weighting function from PML dataset during the training, a transferable calibrator is designed, which takes the histogram of positive samples' confidences as input, and outputs the optimal curriculum weighting for training. Prototype alignment regularization is also proposed to promote the model's performance. 
  Experiments conducted on Pascal VOC, MS-COCO, NUS-WIDE and CUB have verified that our method outperforms existing state-of-the-art PML methods.
\end{abstract}

\begin{links}
 \link{Code}{https://github.com/lee-plus-plus/PML-CD}
\end{links}

%

\section{Introduction}

  \begin{figure*}[t]
    \centering

    \subfigure[Existing weighting functions.]{
      \label{fig:weighting-functions}
      \includegraphics[width=0.24\textwidth]{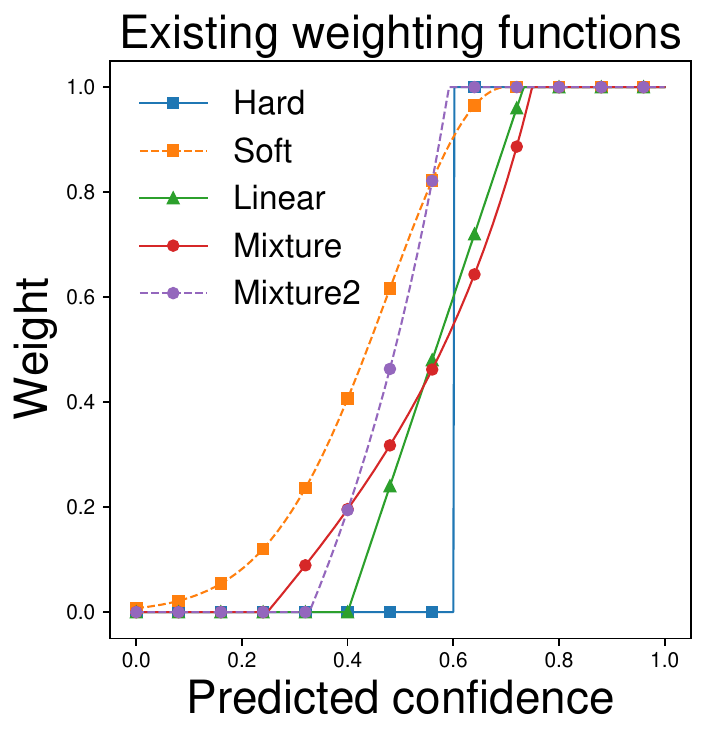}
    }
    \subfigure[Histogram of the predict confidence.]{
      \label{fig:histogram-of-confidences}
      \includegraphics[width=0.47\textwidth]{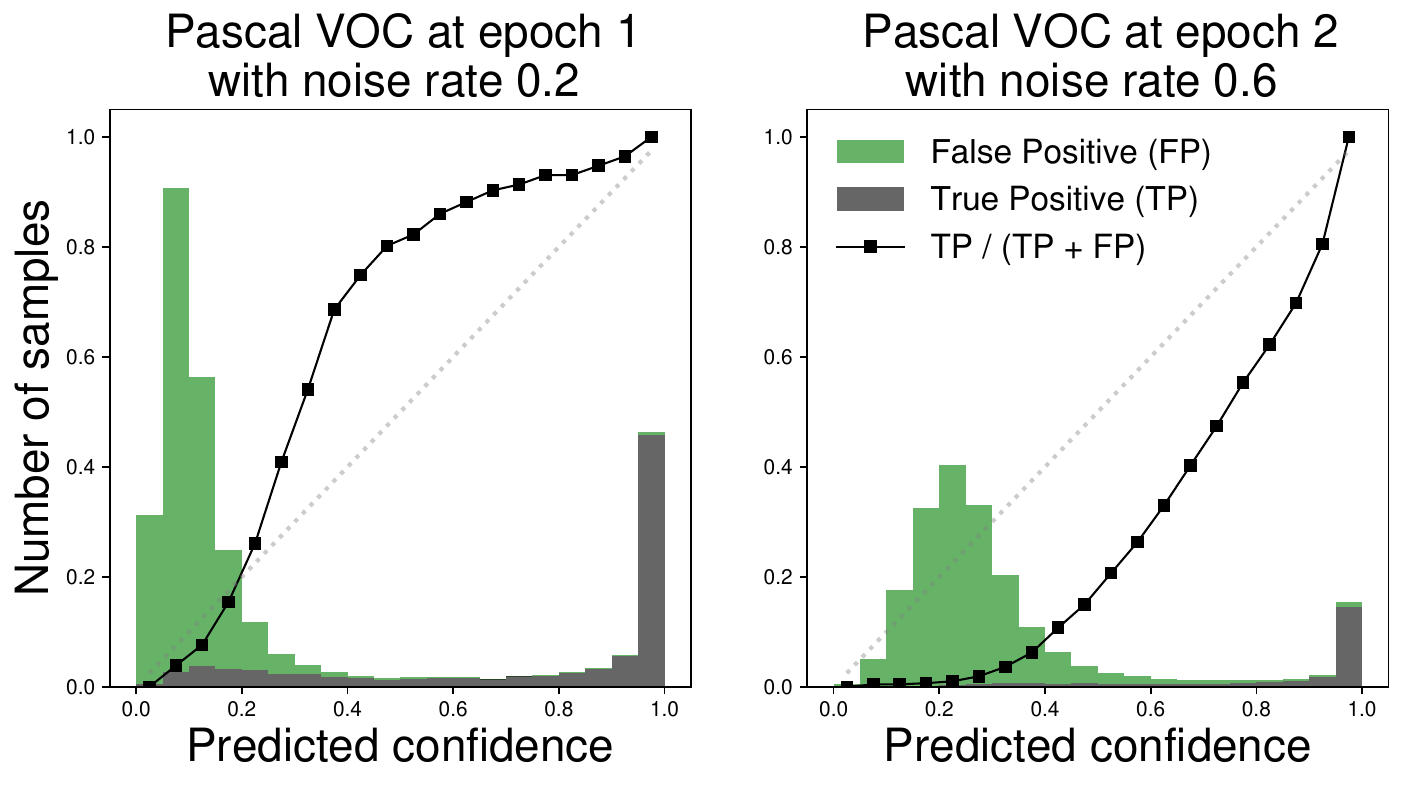}
    }
    \subfigure[Calibrator predict on unseen dataset with high precision.]{
      \label{fig:calibrator-prediction}
      \includegraphics[width=0.24\textwidth]{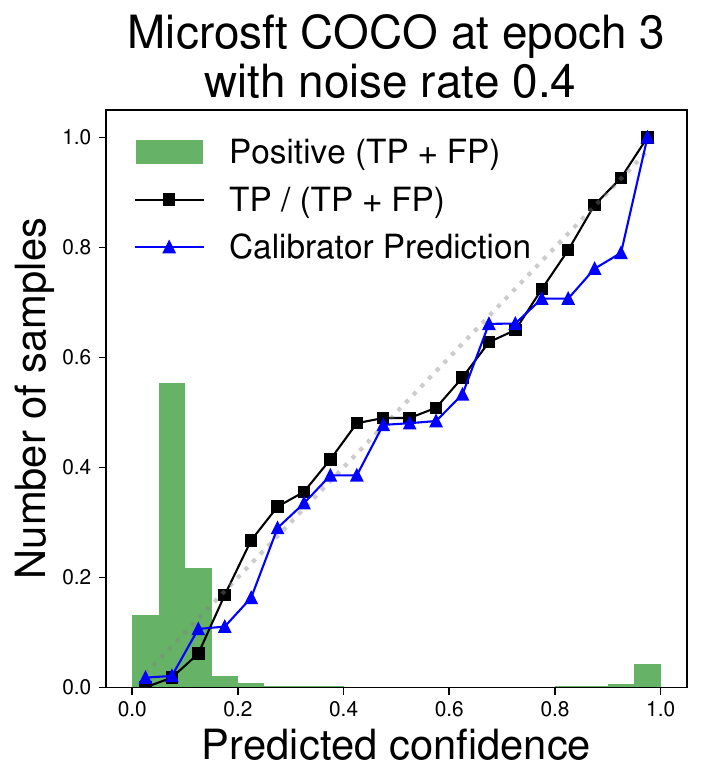}
    }
    \caption{Illustration of Calibrated Disambiguation. (a) shows the weighting function proposed in previous curriculum learning approaches, whose shape are heuristically designed; (b) is two histograms of predicted confidences on clean partial labels and corrupted partial samples. The optimal weighting curve, which is equivalent to the calibration curve, can be derived from the frequency statistics if the ground-truth information is given; (c) shows that the optimal weighting curve can be precisely predicted according to the histogram of predicted confidences on partial labels, perceive a \textbf{general relations between confidences histogram and calibration curve} according to the memorization effect. }
    \label{fig:figure1}
  \end{figure*}

  In recent years, multi-label learning has become a popular research topic, whose goal is to predict whether each category of label is relevant to a sample \cite{EmergingMLL}. Given the fact that precisely annotated datasets are often difficult to be obtained due to high labor cost, crowd-sourcing platforms become common adopted for data collection, irrelevant labels are assigned by non-expert annotators, resulting the samples to be over-annotated, therefore leading to the problem of Partial Multi-label Learning (PML) \cite{fPML}. 
  PML works with the problem that each instance is associated with a candidate label set, which contains not only all the relevant labels but also some irrelevant labels. Several PML methods have been proposed to disambiguate corrupted labels and learn robust model, achieving remarkable performance on PML table datasets \cite{PML-NI,PML-LD,PLAIN}. Inspired by the success of deep neural networks (DNNs), researchers begin to explore the potentials of designing PML methods with DNNs, particularly on PML image classification \cite{PML-CDCR}, yet few methods have been proposed.

  Curriculum learning, which let the model gradually learn from easy knowledge to hard knowledge by controlling the sample's weight during the training \cite{Curriculum-learning-2009}, is learnt to be helpful to disambiguate noisy labels \cite{MentorNet,LL-R}. Therefore, it has been adopted by the majority of existing PML image classification methods \cite{PML-CDCR,UNM}. 
  In curriculum-based PML methods, the weighting function, also be named as the curriculum, is crucial for disambiguation. 
  Among these proposed methods, however, although abundant weighting functions have been proposed (see Figure \ref{fig:weighting-functions}) \cite{Survey-curriculum-learning}, they are heuristically designed, e.g., hard threshold \cite{PML-CDCR}, soft threshold \cite{SoftMatch}, and temperature annealing \cite{Dash}, and therefore provide limited explanation for the criteria of designing curriculum, leading the evaluation or comparison between curriculum impossible.  

  Besides, it is noticed that the suitable curriculum should vary on different dataset, different noise rate and different training epoch. Figure \ref{fig:histogram-of-confidences} provides such an example, showing that the best threshold to separate the true positive and false positive samples according to their predicted confidences vary. Therefore, a proper curriculum weighting function should be adaptive to the feedback of the training. However, existing curriculum-based PML approaches fail to achieve this \cite{PML-CDCR,UNM}. 

  Motivated by the aforementioned limitations, we propose an approach named Calibrated Disambiguation for Partial Multi-label Learning (PML-CD), with a new perspective towards curriculum-based disambiguation by calibration. As it is shown in Figure \ref{fig:histogram-of-confidences}, given the histogram of the confidences of candidate samples, the aim of curriculum weighting is to separate them into clean samples and corrupted samples, assigning high weights for the former and low weights for the latter. 
  Given the fact that samples with same confidence are indiscriminate during the disambiguation, we prove that the optimal mapping curve from samples' confidence to sample's weight should equals to the $TP / (TP + FP)$ curve, which is named as the optimal weighting function. By applying the optimal weighting function during the curriculum-based disambiguation, the potentiality of the memorization effect can be fully mined, providing an upper bound for curriculum-based disambiguation. Figure \ref{fig:calibration} shows that the optimal weighting function is equivalent to the calibration on sample's confidences. 

  In PML training, however, the statistics of whether a positive sample belongs to true positive or false positive, is unknown. To predict the optimal weighting function when training on PML dataset, we designed a transferable calibrator to learn the relationship between the distribution of positive samples' confidence, and the optimal weighting function. A data-driven approach is adopted to train the calibrator, which avoids making assumptions on the distribution of confidences. Figure \ref{fig:calibrator-prediction} shows that our pre-trained calibrator can give fine-grained prediction for optimal weighting curve on unseen dataset. 

  To enhance the robustness of our model, prototype alignment regularization is also proposed, which helps the model learn better multi-label representation and perform better.

  In summary, our contributions are listed as follows:

  \begin{enumerate}

    \item An analysis for the optimal weighting function is provided. which demonstrates the essence of weighting in curriculum-based disambiguation is confidence calibration. 

    \item An adaptive and transferable calibrator is developed, which is based on the common patterns observed between the confidence distribution histograms of candidate samples and the ratio of clean samples.

    \item A prototype alignment regularization loss is proposed, which can help multi-label encoder produce not only separable features, but also discriminative features, therefore leading to robust performance.

  \end{enumerate}

\section{Related Work}

\subsection{Multi-label Image Classification}

  In recent years, multi-label image classification has drawn increasing attention, since it is more reasonable for each image to contain multiple relevant labels\cite{GAELLE}. To fully exploit the label correlations, convolutional neural networks (CNNs) with new architectures have been proposed according to various perspectives, including recurrent neural networks \cite{cnn-rnn}, graph convolutional networks \cite{ML-GCN,ML-AGCN}, graph propagation \cite{Scene-aware}, and Transformer architecture \cite{Query2Label,C-Tran}.


\subsection{Partial Multi-label Learning}
  
  Partial Multi-label Learning (PML) considers a situation where each instance is associated with a set of candidate labels, while only part of them are the ground-truth labels. 
  Some existing work utilizes global and local information to guide disambiguation, following the manifold assumption and the label correlation assumption \cite{PML-LD,DRAMA,PML-SALC}. Other existing works, following the low-rank decomposition assumption, prefer to separate ground truth and noise from corrupted data by applying low-rank constraints while optimizing \cite{PML-LRS,PML-NI}. 
  Lately proposed PML methods usually incorporate both the manifold and low-rank assumptions in a delicate way \cite{GLADE,PAKS,GLC,NLR}. 

  In recent years, researchers began exploring the compatibility of PML with DNNs and investigated PML image classification. Current image PML methods are primarily based on memory effects \cite{a-closer-look}, and apply curriculum-guided disambiguation through small-loss criteria \cite{PML-CDCR,UNM}.

\subsection{Curriculum Learning}


  The concept of curriculum learning is inspired by the human learning process, where individuals first acquire simple and general knowledge, then gradually learns more complex and specialized knowledge and thereby achieving well understanding \cite{Curriculum-learning-2009}. 
  By leveraging the memorization effect \cite{a-closer-look} and employing training sample loss as the difficulty measurer \cite{Survey-curriculum-learning}, noisy label learning approaches can progressively disambiguate labels through curriculum learning strategies \cite{MentorNet}. 


  Based on the small-loss criteria, a series of weighting functions that map samples' loss to their weights have been proposed in curriculum learning \cite{Survey-curriculum-learning}, with various forms such as hard \cite{kumar2010self}, linear \cite{Easy-samples-first}, logarithmic \cite{Easy-samples-first}, Gaussian \cite{SoftMatch}, and etc. Since the most appropriate curriculum  varies on different training conditions, the weighting function should be adaptively adjusted by the feedback during the training. 
  Existing approaches have proposed various curriculum adaptation strategies, by introducing the confidence histograms \cite{CLC}, the mean and variance of predicted confidence \cite{SoftMatch}, quantized logical differences \cite{DynaCor}, loss rankings \cite{UNM}, and etc. Corresponding parameter updating schemes have also been designed in different approaches, including fitting two Gaussian distribution \cite{CLC}, temperature annealing \cite{Dash}, and etc.

  Overall, \textbf{designing adaptive weighting strategies heuristically is challenging for three main reasons}: 1) a mathematical formulation for optimal curriculum lacks; 2) certain feature construction approaches, such as ranking, are inherently suboptimal for precise inference; 3) some predictors employ generative models that rely on assumptions, such as the confidence levels of noisy and clean samples following two separable normal distributions \cite{CLC}, which may not always be applicable.

  In addition to re-weighting, some approaches explore how to re-label corrupted labels in a curriculum-based manner \cite{Curriculum-learning-2009,LL-R}, or combine re-weighting and re-labeling together \cite{RoCL}. It is also referred to as re-labeling or label correction. The main challenge with re-labeling lies in the possibility of incorrectly correcting difficult but accurate labels in the early stages of training, which could even damage further learning \cite{LL-R}. 

\section{Preliminaries}
\label{sec:Preliminaries}

  We first provide a detailed explanation of the notations used in the paper. The multi-label training dataset is denoted as \(\mathcal{D} = \{(x_i, y_i) \mid i=1,\dots, N\}\), where \(N\) is the number of samples of the training set, \(x_i \in \mathbb{R}^{W\times H\times3}\) is the image of the \(i\)-th sample, and \(y_i = [y_i^1, \dots, y_i^C]^T \in \{0, 1\}^C\) denotes the ground truth labels for each category of the \(i\)-th sample, with a total of \(C\) categories. In PML, since labels are not precisely annotated, only a corrupted dataset \(\tilde{\mathcal{D}} = \{(x_i, \tilde y_i) \mid i=1,\dots,n\}\) is available, whose candidate labels \(\tilde y_i = [y_{i1}, \dots, y_{ic}]^T \in \{0, 1\}^C\) for the \(i\)-th sample contains not only relevant labels corresponding to \(y_i\), but also additional noise. For convenience, we denote \(\{\cdot\}\) in this paper to succinctly represent a collection of certain variable of every sample on the training set, e.g., \(\{x_i\}\).

  \begin{figure}[tb]
    \centering    {\includegraphics[width=1.0\linewidth]{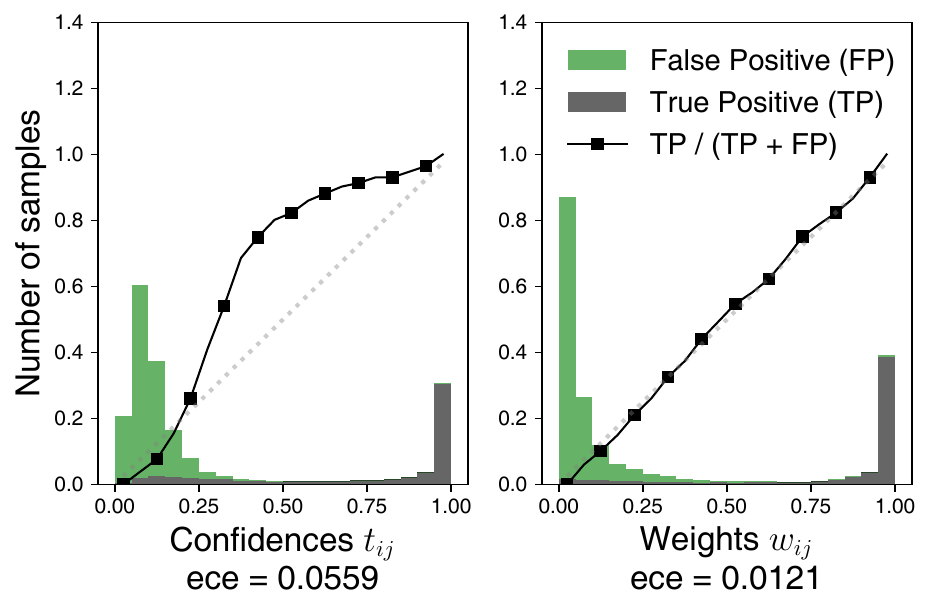}}
  \caption{An example of optimal curriculum weighting with histograms of model's predicted confidences (left) and weights (right) by assigning \(w_{ij} = P(y_{ij} = \tilde y_{ij} \mid t_{ij})\). It is shown that weighting is equivalent to calibrating model's predicted confidences. The expected calibration error (ECE) of confidences and weights are presented. \textbf{}}
  \label{fig:calibration}
  \end{figure}

  Given a model \(f: \mathbb{R}^{W \times H \times 3} \rightarrow [0, 1]^{C}\), the confidence predictions \(t_i = f(x_i) \in [0, 1]^{C}\) for each sample on training set can be collected.
  According to the memorization effect, when a model learns on a noisy dataset, it tends to first memorize the clean samples. It suggests that the small-loss criteria to filter reliable samples and assign higher weights to samples with lower loss during the training. Such disambiguation procedure is also known as curriculum learning, which can be formulated as
  \begin{equation}
    \ell_\text{curr}(\hat y_{ij}, \tilde y_{ij}) = w_{ij} \ell_\text{BCE}(\hat y_{ij}, \tilde y_{ij}),
  \end{equation}
  where $\hat y_{ij} = f_j(x_i)$ is the model's predicted confidence for $i$-th sample and $j$-th class, and $\ell_\text{BCE}(\hat y, \tilde y) = -\left(\tilde y \log(\hat y) + (1 - \tilde y) \log(1 - \hat y)\right)$ is the binary cross entropy loss.
  Given a predicted score $t$, we hope to assign weight $w(t)$ to samples to better train the model. 
  It is expected to assign as much weight as possible to clean samples, and as little weight as possible to corrupted samples. which can be evaluated by
  \begin{equation}
  \begin{aligned}
    \ell_\text{BCE}\left(w(t), \mathbb{I}[y = \tilde y] \right) 
    & = \underbrace{- \mathbb{I}[y = \tilde y] \cdot \log (w(t))}_\text{maximize clean samples' weights}
      + \\ & \underbrace{(-\mathbb{I}[y \neq \tilde y]) \cdot \log (1 - w(t))}_\text{minimize corrupted samples' weights},
  \end{aligned}
  \end{equation}
  and the \textit{optimal} weighting function $w^*(t)$ can be induced as
  \begin{equation}
  \begin{aligned}
    w^{*}(t) = & \arg \min_{w(t)} \ \mathbb{E}_{y \mid t} \left[ \ell_\text{BCE}(w(t),\mathbb{I}[y=\tilde y]) \right] \\
    = \ & P(y=\tilde y \mid t).
  \end{aligned}
  \end{equation}
  Since other baseline methods try to model $w^*(t)$ in a binarization way, e.g., $w(t) = \mathbb{I}[t \geq \eta]$, our method provides theoretical upper bound. 
  Therefore, the optimal curriculum weighting function is to model the posterior probability $P(y_{ij} = \tilde y_{ij} \mid t_{ij})$ as
  \begin{equation}
  \label{eq:optimal-weight}
    \ell^*_\text{curr}(\hat y_{ij}, \tilde y_{ij}, y_{ij}) 
    = P(y_{ij} = \tilde y_{ij} \mid t_{ij}) \ell_\text{BCE}(\hat y_{ij}, \tilde y_{ij}).
  \end{equation}

  To better understand the meaning of Equation \ref{eq:optimal-weight}, an example of optimal curriculum weighting is given in Figure \ref{fig:calibration}. 
  In summary, our analysis below provides three insights:

  \begin{enumerate}

    \item In curriculum learning based disambiguation schemes, \textbf{the essence of weighting is the calibration of model's predicted confidence}. 

    \item The optimal weighting function $P(y_{ij} = \tilde y_{ij} \mid t_{ij})$ provides a \textbf{theoretically upper bounds for small-loss guided disambiguation}. 

    \item The \textbf{quality of weighing function can be evaluated according to the quality of calibration}, via metrics like expected calibration error (ECE). 

  \end{enumerate}

\section{Methods}

  \begin{figure*}[t]
    \centering    {\includegraphics[width=0.9\textwidth]{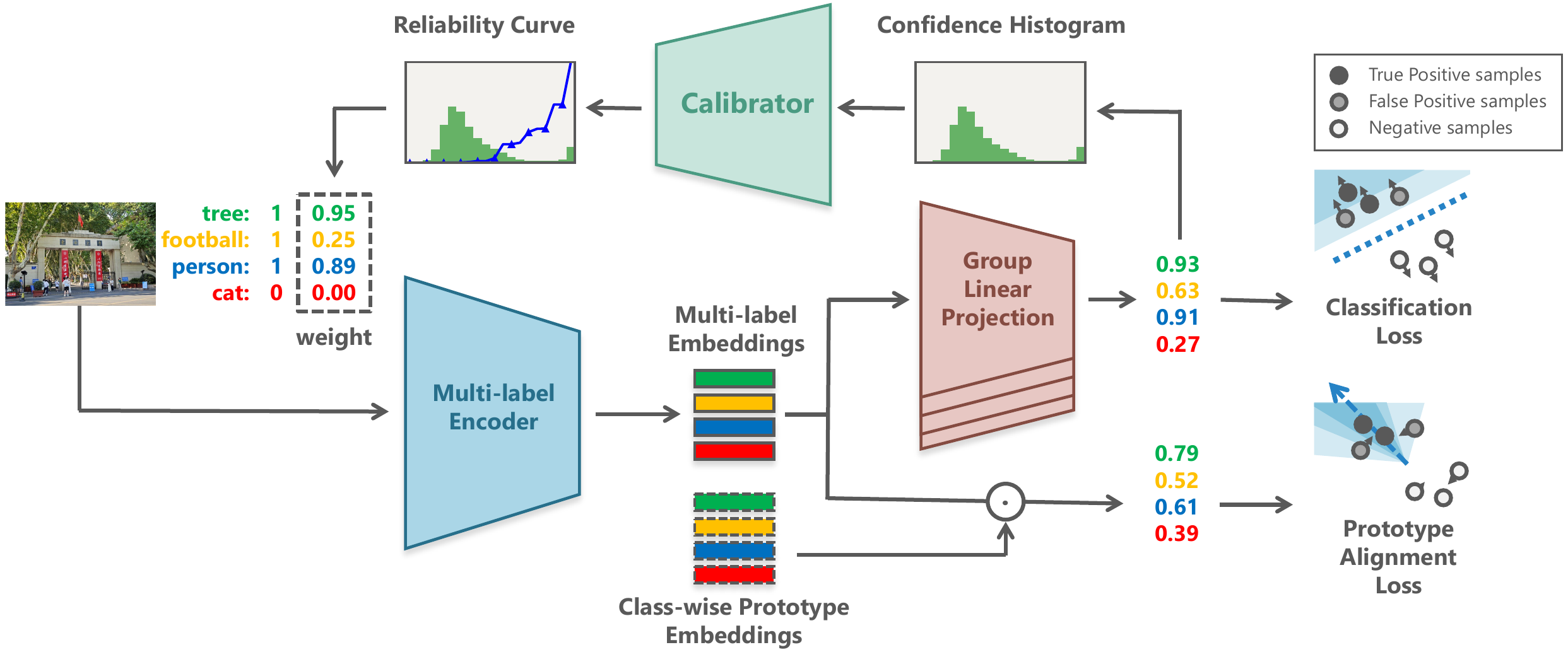}}
    \caption{Illustration of the proposed model PML-CD. During the training, the model is updated by both classification loss and prototype alignment loss. Predicted confidences on training set are collected and are fed into the calibrator to generate reliability of candidate labels, which guides the curriculum disambiguation training in the next epoch. }
  \label{fig:framework}
  \end{figure*}

  In this section, we first introduce the overall architecture and training process of our proposed model PML-CD, which is also presented in Figure \ref{fig:framework}. Design proposals and implementation details of each module will be introduced later.

  \textbf{Two phase training}. 
  The training progress of our PML-CD model is divided into two phases. The first phase is the warm-up phase, where the model is trained via Binary Cross-Entropy (BCE) loss on raw training set until early stopping \cite{EarlyStopping}. This phase aims to initialize the model's random parameters to a reasonable state and to produce reasonable multi-label representations. 
  The second phase is the disambiguation phase. In this phase, we collect the model's predicted confidence on the training set, and apply the calibrator to estimate the reliability of candidate labels, then apply the reliability score as the weights in a curriculum learning. The model is robustly trained via weighted BCE loss along with the prototype alignment regularization until early stopping. 

  \textbf{Multi-label encoder}. 
  Current researches in multi-label learning suggests that it is more appropriate to learn a representation for each categories in an image. In multi-label images, objects of different classes are often situated in distinct regions of the image and can sometimes be quite small. Therefore, compressing all regional features into one feature vector through adaptive pooling is suboptimal \cite{,Query2Label,GAELLE}. Applying attention mechanisms on regional features enables the model to learn which regional features are relevant to each class, thus helping the model to learn more consistent feature representations \cite{,Query2Label,ML-Decoder}. Following the architecture of ML-Decoder \cite{ML-Decoder}, we first extract regional features from the image using pre-trained CNN backbone, and then pass them into a Transformer with $C$ query vectors, therefore outputs the multi-label representations of an image $z \in \mathbb{R}^{C \times D}$. 

  \textbf{Group linear projection}. 
  To produce confidence predictions, the Group Linear Projection \cite{ML-Decoder} is adopted, where the representation in each class is passed through its corresponding linear layer. 

\subsection{Calibrator}

  Following the above analysis that \textbf{the essence of weighting is calibration}, unlike previous methods that heuristically formulating weighting function, we formulate our calibrator from the analysis of the optimal calibrator. Given the information of the ground-truth labels $\{y_i\}$, the optimal calibrator can be derived from the confidences on true partial labels and false partial labels in a histogram way. 
  Given the predicted confidence on partial labels $\{t_{ij} \mid \tilde y_{ij} = 1\}$, we split the partial confidences
    $\{t_{ij}\}^\text{P} = \{t_{ij} \mid \tilde y_{ij} = 1\}$ 
  into two groups, the True Positive (TP) group 
    $\{t_{ij}\}^\text{TP} = \{t_{ij} \mid \tilde y_{ij} = 1, y_{ij} = 1\}$ 
  and the False Positive (FP) group
    $\{t_{ij}\}^\text{FP} = \{t_{ij} \mid \tilde y_{ij} = 1, y_{ij} = 0\}$. 
  We denote the number of bins $K = 20$, confidences in each bin $\{t_{ij}\}^{[k]} = \{ (k-1)/K \leq t_{ij} < k/K\}$, and then the frequency distribution of TP and FP confidences are collected as
  \begin{equation}
  \label{eq:frequency}
  \begin{aligned}
    & q^\text{P}_k = \frac{ \lvert \{t_{ij}\}^{[k]} \cap \{t_{ij}\}^\text{P} \rvert }{ \lvert \{t_{ij}\}^\text{P} \rvert }, \\
    & q^\text{FP}_k = \frac{ \lvert \{t_{ij}\}^{[k]} \cap \{t_{ij}\}^\text{FP} \rvert }{ \lvert \{t_{ij}\}^\text{P} \rvert }, \\
    & q^\text{TP}_k = \frac{ \lvert \{t_{ij}\}^{[k]} \cap \{t_{ij}\}^\text{TP} \rvert }{ \lvert \{t_{ij}\}^\text{P} \rvert }. \\
  \end{aligned}  
  \end{equation}
  Therefore, the reliability histogram can be derived as
  \begin{equation}
  \label{eq:reliablity}
    r_k = q^\text{TP}_k / (q^\text{TP}_k + q^\text{FP}_k).
  \end{equation}
  In the end, we can export the confidence-reliability relationship between $t_i$ and $P(y_{ij} = \tilde y_{ij} \mid t_{ij})$ via applying interpolation on reliability histogram 
  \begin{equation}
  \label{eq:interpolation}
  \begin{aligned}
    & P(y_{ij} = \tilde y_{ij} \mid t_{ij}) = (\frac{k+1}{K} - t_{ij}) r_k + (t_{ij} - \frac{k}{K}) r_{k+1}, \\
    s.t. &  \ \frac{k}{K} \leq t_{ij} < \frac{k+1}{K}.
  \end{aligned}
  \end{equation}
  To produce precise calibration, the confidence-reliability relationship can be individually accomplished on each class. 

  It has been witnessed that the histogram of predicted confidence is highly relevant to the ground-truth confidence-reliability relationship during the training \cite{CLC,UNM}. 
  Reasonable attempts that fit the TP and FP confidences via two Gaussian distribution has been made \cite{CLC}. However, according to our observation, it can hardly be adopted to general situations where the histogram of TP and FP confidences are skewed or are highly overlapping. 

  To overcome this issue, a \textit{data-driven approach} is adopted. We model our calibrator \( g: [0, 1]^K \rightarrow [0, 1]^K \) with the frequency of confidences in each bin \(\bigl(q^\text{P}\bigr)_{k \in K}\) as input, and the confidence-reliability relationship $\bigl(q^\text{TP}_k / (q^\text{TP}_k + q^\text{FP}_k )\bigr)_{k \in K}$ as output. The structure of the calibrator is designed as a 2-layer Multi-layer Perceptron with the activation of LeakyReLU. 
  To enable the calibrator to learn the relationship, we collect information of $\{q^\text{P}, q^\text{TP}, q^\text{FP} \}$ during the training of baseline model on every epoch at the disambiguation phase. The baseline model is implemented with the same architecture as PML-CD but adopts a trivial disambiguation strategy $P(y_{ij} = \tilde y_{ij} \mid t_{ij}) = t_{ij}$ adopted. In order to enable our calibrator be adaptive to different situation, the training data is collected from multiple training procedures on a dataset with different noise rate. After pre-training, the calibrator can be run on any dataset with any noise rate by loading parameters, and therefore serving as a plugin. 
  Since our calibrator avoids making assumptions on the distribution of TP and FP confidences, and learns from data with both low noise and high noise, it can adapt to various scenario. 


\subsection{Prototype Alignment Regularization}

  It is common for multi-label DNN methods to first encode the multi-label representation, then produce the confidence through a \textit{Linear Projection} layer for each class. Given the multi-label representation of \(i\)-th sample on \(j\)-th class $z_{ij} \in \mathbb{R}^{D}$, and the parameter of \(j\)-th class linear projection layer $\theta_j \in \mathbb{R}^D, b_j \in \mathbb{R}$, the confidence $\hat y_{ij}^\text{clf}$ is predicted as
  \begin{equation}
    \hat y_{ij}^\text{clf} = \theta_j^T z_{ij} + b_j.
  \end{equation}
  Therefore, the classification loss 
  \begin{equation}
    \mathcal{L}_\text{clf} = \sum_{i \in N} \sum_{j \in C} \ell_\text{curr}(\hat y_{ij}^\text{clf}, \tilde y_{ij})
  \end{equation}
  can be viewed as seeking an optimal hyperplane to separate the representations of positive and negative examples for each class. However, the decision boundary learned through linear projection can be easily disrupted by noisy labels, therefore lacking robustness \cite{CTRR}. On the other hand, another widely applied way to produce confidence via multi-label representations is \textit{Prototype Alignment}, which is reckoned to be robust in noisy label learning and is less prone to be over-fitting. Given the prototype representation of \(j\)-th class $\bar z_{j} \in \mathbb{R}^{D}$, the confidence $\hat y_{ij}^\text{proto}$ is predicted as
  \begin{equation}
    \hat y_{ij}^\text{proto} = \frac{\langle z_{ij}, \bar z_{j} \rangle}{ \|z_{ij} \| \cdot \| \bar z_{j} \|}.
  \end{equation}
  Given the prototype alignment loss 
  \begin{equation}
    \mathcal{L}_\text{proto} = \sum_{i \in N} \sum_{j \in C} \ell_\text{curr}(\hat y_{ij}^\text{proto}, \tilde y_{ij}),
  \end{equation}
  unlike maximizing the distance of representations from the decision hyperplane, the prototype alignment method compute the cosine similarity between sample's representation and the corresponding class prototype. It has been widely applied in partial label learning to guide disambiguation and provide more reliable label guesses for uncertain samples \cite{PiCO, PiCO-plus, CEL}. However, this projection method struggles to achieve fine-grained fitting, resulting in suboptimal prediction performance.

  Taking the impact on the encoder into consideration, optimization on linear projection score
    $\ell_\text{curr}(\hat y_{ij}^\text{clf}, \tilde y_{ij})$
  lets the embeddings of positive and negative samples linear separable, while optimization on prototype alignment score
    $\ell_\text{curr}(\hat y_{ij}^\text{proto}, \tilde y_{ij})$
  lets the embeddings of positive and negative be clustered around their prototype embeddings. By applying both of them, 
  a representation which is not only separable, but also discriminative, is encouraged \cite{CenterLoss}. 

  Above all, in warm-up phase, the training loss is
  \begin{equation}
  \label{eq:loss-warmup}
    \mathcal{L}_\text{warmup} = \mathcal{L}_\text{clf},
  \end{equation}
  and training loss in disambiguation phase is defined as
  \begin{equation}
  \label{loss-disambiguation}
    \mathcal{L}_\text{disam} = \mathcal{L}_\text{clf} + \mathcal{L}_\text{proto}.
  \end{equation}
  During the training, we allow the prototype representations be updated via gradient, therefore no explicit updating strategy is required. 
  Since the multi-label representation is not well initialized during the warm-up training, making their prototypes meaningless, the prototype alignment regularization is only applied during the disambiguation phase.

  \begin{algorithm}[tb]
    \textbf{Input}: 
      partial multi-label learning dataset $\{(x_i, \tilde y_i)\}$,
      \\
    \textbf{Parameter}: 
      pre-trained calibrator \(g\),
      learning rate $\eta$
      \\
    \textbf{Output}: 
      network $f(\cdot \mid \theta)$ for PML
    \begin{algorithmic}[1] 
      \REPEAT
        \STATE Update network parameters $\theta \leftarrow \theta - \eta \Delta \mathcal{L}_\text{warmup}$
      \UNTIL{early stopping}
      \STATE Initialize $w_{ij} \leftarrow 1$
      \REPEAT
        \STATE Collect predicted confidences $\{t_{ij} \} \leftarrow f(\{x_i\})$
        \STATE Compute $q^\text{P}$ via Equation. \ref{eq:frequency}
        \STATE Predict reliability histogram $r$ via Equation. \ref{eq:reliablity}
        \STATE Update curriculum weights $w_{ij} = P(y_{ij} = \tilde y_{ij} \mid t_{ij})$ via Equation. \ref{eq:interpolation}, if $\tilde y_{ij} = 1$
        \STATE Update network parameters $\theta \leftarrow \theta - \eta \Delta \mathcal{L}_\text{disam}$
      \UNTIL{early stopping}
    \end{algorithmic}
    \caption{Algorithm of PML-CD}
    \label{alg:algorithm}
  \end{algorithm}

\section{Experiments}

\subsection{Settings}

  \textbf{Datasets}. To validate our proposed approach, we conducted experiments on several commonly used real-world multi-label image datasets, including MS-COCO 2014 \cite{COCO}, Pascal VOC 2007 \cite{VOC}, NUS-WIDE \cite{NUS-WIDE}, and CUB 200 \cite{CUB}. 
  Specifically,  we took the 312 binary attributes instead of the bird category as predicted labels, therefore making it a multi-label dataset. To create PML datasets from multi-label datasets, we injected random noise into the label matrices by randomly flipping the zero values in the original label matrix with a certain probability, denoted as noise rate $\rho$. 

  \textbf{Comparison Methods}. We compare our PML-CD model with three groups of methods. 

  \begin{table*}[tb]
    \renewcommand{\arraystretch}{1.0}
    \small
    \setlength\tabcolsep{3.6pt}
    \centering
    \begin{tabular}{lcccccccccccccccc}
    \toprule
      & \multicolumn{4}{c}{VOC} & \multicolumn{4}{c}{COCO} & \multicolumn{4}{c}{NUS-WIDE} & \multicolumn{4}{c}{CUB} \\
      \cmidrule(r){2-5}\cmidrule(r){6-9}\cmidrule(r){10-13}\cmidrule(r){14-17}
      Method & 20\% & 40\% & 60\% & 80\% & 20\% & 40\% & 60\% & 80\% & 20\% & 40\% & 60\% & 80\% & 20\% & 40\% & 60\% & 80\% \\
      \midrule
      PML-NI     & $62.57$&$35.37$&$20.97$&$13.84$   & $60.14$&$53.41$&$43.02$&$24.52$ & $21.26$&$16.39$&$12.02$&$\hphantom{0}7.89$  & $12.92$&$12.04$&$11.42$&$10.96$ \\
      PML-LRS    & $79.96$&$63.45$&$32.86$&$16.66$   & $61.56$&$53.80$&$42.85$&$24.23$ & $21.52$&$16.23$&$11.85$&$\hphantom{0}7.76$  & $14.56$&$12.91$&$11.83$&$11.19$ \\
      PLAIN      & $80.76$&$72.45$&$54.59$&$27.24$   & $60.13$&$53.71$&$44.99$&$21.56$ & $19.22$&$17.85$&$13.72$&$\hphantom{0}7.44$  & $14.98$&$11.38$&$10.55$&$10.37$ \\
      \midrule
      BCE        & $87.73$&$84.26$&$75.27$&$48.86$   & $73.66$&$69.62$&$63.52$&$50.47$ & $50.81$&$45.78$&$37.90$&$27.06$  & $29.66$&$27.01$&$22.19$&$17.39$ \\
      ASL        & $88.63$&$84.43$&$77.23$&$52.95$   & $73.19$&$68.53$&$62.56$&$50.46$ & $46.27$&$37.86$&$31.67$&$21.97$  & $28.58$&$25.75$&$22.66$&$17.35$ \\
      MLDecoder  & $87.94$&$85.49$&$80.41$&$64.46$   & $74.29$&$70.35$&$68.55$&$60.55$ & $50.92$&$46.11$&$40.01$&$30.83$  & $29.72$&$27.63$&$24.99$&$21.13$ \\
      \midrule
      UNM        & $87.99$&$86.58$&$79.94$&$44.07$   & $65.56$&$63.51$&$61.68$&$27.87$ & $37.83$&$34.57$&$29.62$&$12.04$  & $23.93$&$19.73$&$18.42$&$11.66$ \\
      PML-CDCR   & $88.56$&$86.87$&$69.26$&$38.31$  & $72.44$&$69.90$&$66.88$&$42.17$ & $42.91$&$39.96$&$31.99$&$22.37$  & $26.67$&$24.63$&$23.05$&$16.38$ \\
      LL-R       &\cellcolor{gray!30}$88.99$&$82.87$&$72.36$&$40.32$   & $74.35$&$67.79$&$61.47$&$41.28$ & $43.75$&$42.09$&$30.56$&$24.76$  & $29.03$&$26.47$&$22.54$&$18.62$ \\
      PML-CD     & $88.83$&$\cellcolor{gray!30}87.12$&\cellcolor{gray!30}$83.87$&\cellcolor{gray!30}$71.70$  & \cellcolor{gray!30}$75.26$&\cellcolor{gray!30}$74.65$&\cellcolor{gray!30}$69.88$&\cellcolor{gray!30}$62.10$ & \cellcolor{gray!30}$51.30$&\cellcolor{gray!30}$46.11$&\cellcolor{gray!30}$40.41$&\cellcolor{gray!30}$32.10$  & \cellcolor{gray!30}$30.52$&\cellcolor{gray!30}$27.87$&\cellcolor{gray!30}$25.42$&\cellcolor{gray!30}$21.69$ \\
    \bottomrule
    \end{tabular}%
    \caption{Comparison of our model PML-CD with baseline methods, with noise rate $20\%$, $40\%$, $60\%$ and $80\%$. mAP($\%$) is adopted as metric. The best performances on each column are highlighted with background color. }
    \label{tab:main-result}%
  \end{table*}

  \begin{itemize}

    \item \textbf{Multi-label learning methods}. \textbf{BCE} is a standard multi-label classification model trained with binary cross entropy loss. \textbf{ASL} \cite{ASL} is a multi-label learning method that addresses the issue of imbalance between positive and negative labels. \textbf{ML-Decoder} \cite{ML-Decoder} re-designs the decoder architecture and introduces a group-decoding scheme to provide efficiency. These methods do not specifically address noise disambiguation, and have achieved state-of-the-art results on many multi-label image classification datasets.

    \item \textbf{Traditional PML methods}. \textbf{PML-NI} \cite{PML-NI} applies low-rank constraints to decompose the PML label matrix, while \textbf{PML-LRS} \cite{PML-LRS} decomposes the projection matrix of linear regression model. \textbf{PLAIN} \cite{PLAIN} disambiguates gradually with instance similarity and label correlations. These methods are shallow methods that are not specifically designed for image classification. Therefore, we use ResNet-101 \cite{ResNet} pre-trained on ImageNet to extract feature embeddings to make them applicable.

    \item \textbf{Curriculum-based PML methods}. \textbf{LL-R} \cite{LL-R} ignores training samples with high loss during the training. \textbf{PML-CDCR} \cite{PML-CDCR} trains on strongly-augmented images and selecting reliable samples on weakly-augmented images with a loss threshold. \textbf{UNM} \cite{UNM} applies cyclical learning rate to transform the model from over-fitting status back to under-fitting status, therefore providing more reliable disambiguation during the training. These methods adopt a curriculum disambiguation framework similar to ours.

  \end{itemize}

  \textbf{Setting and evaluation}. Our experiments are run on GeForce RTX 4090 with PyTorch 1.13.1. We employ ResNet-101 \cite{ResNet} pre-trained on ImageNet as the backbone, and the detailed setting for all experiments are described as follows. 

  During the training, all images are resized to $224 \times 224$, and strong data augmentation are applied for training set, including horizontal flip, RandAugment \cite{Randaugment} and Cutout. The same data processing is adopted for almost all comparison methods. We train the model with the following optimization setting: an Adam Optimizer with fixed learning rate $1 \times 10^{-4}$ for VOC and CUB, $1 \times 10^{-5}$ for COCO and NUS-WIDE; a weight decay of $5 \times 10^{-5}$. The size of mini-batch is set to be $32$. Early stopping is applied for all comparison methods, since noisy labels learning is highly dependent on it \cite{EarlyStopping}.
  \textbf{Mean Average Precision} (mAP) is applied to evaluate models' performance. 

  \textbf{Calibrator pre-training}. Our calibrator is pre-trained with the following procedure. First, the baseline model without prototype alignment regularization loss and with an identity weighting function is trained on VOC dataset, with noise rate ranging from $10\%, 20\%, \dots, 90\%$. 
  The partial label matrix, ground-truth label matrix, and the confidence prediction matrix at every epoch are collected during the training. On every epoch, the confidence histogram and the reliability histogram for each class are calculated independently, therefore making \(C\) feature-label pairs $( q^\text{P}, r )$. After the dataset is constructed, the calibrator is trained by minimizing \textbf{Mean Squared Error} (MSE) via an Adam optimizer with a fixed  learning rate $1 \times 10^{-2}$. 

\subsection{Comparison Results}

  Results are reported in Table \ref{tab:main-result}, which shows that our model outperforms other baseline models on $15/16$ conditions, thus achieving state-of-the-art results. PML-CD performs slightly inferior to other baseline methods under low noise condition on VOC, but performs significantly superior to others under high noise conditions, which can be explained as PML-CDCR and LL-R is more suitable on low noise condition since they adopt simpler network architecture, i.e., ResNet only. It is noticed that our model performs well on CUB, which has over 300 fine-grained visual attributes, making it challenging for PML. Our model outperforms other curriculum-based PML baseline models with significant advantages, for the reason that our model adopts the ML-Decoder's multi-layer transformer architecture as the encoder, therefore advantageous in extracting fine-grained visual features. 

  To summarize, our model outperforms other baseline models for $15/16$ times under different training conditions, which shows that our model, together with the pre-trained calibrator, can not only perform well on VOC dataset, but also on unseen dataset including COCO, NUS-WIDE, and CUB. Given the complexity of multi-label encoder can sometimes lead to sub-optimal performance on on easy PML dataset, e.g., VOC with low noise rate, it enables our model to encode fine-grained visual features on images, which is of necessity when training on CUB. 

\subsection{Ablation Study}

  \textbf{Effectiveness of calibrator}. To evaluate whether the pre-trained calibrator can accurately predict the optimal calibration curve on unseen datasets, and therefore calibrate their predicted confidences, we evaluate the expected calibration error (ECE) of predicted confidences and the curriculum weighting, during the disambiguation phase of the training on VOC, COCO datasets. Also, to compare with other weighting policy, we adopt the pre-trained calibrator, the optimal calibrator, and other commonly used curriculum weighting functions for comparison. We denote 
  the \textbf{hard} weighting function 
    $v^\text{hard}(t_{ij}) = \mathbb{I}[t_{ij} \geq 0.6]$, 
  the \textbf{soft} weighting function 
    $v^\text{soft}(t_{ij}) = (t_{ij})^{2}$,
  the \textbf{identity} weighting function 
    $v^\text{identity}(t_{ij}) = t_{ij}$.
  Comparison results are shown in Table \ref{tab:weighting}. 

  \begin{table}[htbp]
    \footnotesize
    \setlength\tabcolsep{5pt}
    \centering
    \begin{tabularx}{0.95\linewidth}{XXXX}
    \toprule
      weighting \newline policy & ECE of \newline confidences \newline  (epoch 1) & ECE of \newline weights \newline (epoch 1) & mAP \newline per class \newline (overall) \\
    \midrule
      hard        & $76.78$ & $92.96$ & $59.79$ \\
      soft        & $76.78$ & $61.68$ & $60.08$ \\
      identity    & $76.78$ & $76.78$ & $59.93$ \\
      calibrator  & $76.78$ & \cellcolor{gray!30}$\hphantom{0}3.74$ & \cellcolor{gray!30}$61.48$ \\
      optimal     & $76.78$ & \cellcolor{gray!50}$\hphantom{0}3.74$ & \cellcolor{gray!50}$66.42$ \\
    \bottomrule
    \end{tabularx}%
    \caption{Effectiveness of weighting policies on MS-COCO with noise rate $80\%$. ECE($\%$) and mAP($\%$) are presented. }
    \label{tab:weighting}%
  \end{table}

  \textbf{Effectiveness of prototype alignment regularization}. To verify whether the prototype alignment regularization (PAR) enhances the performance of PML-CD, we train our model on the VOC and COCO datasets with corresponding settings: 1) applying no PAR; 2) applying PAR at the disambiguation phase; 3) applying PAR at both the warmup phase and disambiguation phase. Comparison results are presented in Table \ref{tab:regularization}. It is shown that prototype alignment regularization helps the model perform better especially in high noise condition. 

  \begin{table}[htbp]
    \footnotesize
    \setlength\tabcolsep{5pt}
    \centering
    \begin{tabular}{cllcccc}
    \toprule
      & & \multicolumn{4}{c}{noise rate} \\
      \cmidrule(r){3-6}
      dataset & $\mathcal{L}_\text{proto}$ & $20\%$ & $40\%$ & $60\%$ & $80\%$  \\
    \midrule
      \multirow{3}{*}{VOC}
      & neither           & \cellcolor{gray!30}$89.28$&$87.93$&$85.05$&$66.91$ \\
      & 2nd-phase only    & $88.77$&\cellcolor{gray!30}$88.12$&\cellcolor{gray!30}$85.07$&\cellcolor{gray!30}$71.70$ \\
      & 1st \& 2nd-phase  & $87.89$&$86.11$&$83.90$&$71.40$ \\
    \midrule
      \multirow{3}{*}{COCO}
      & neither           & \cellcolor{gray!30}$75.54$&$73.86$&$69.78$&$60.13$ \\
      & 2nd-phase only    & $75.26$&\cellcolor{gray!30}$74.65$&\cellcolor{gray!30}$69.88$&$62.10$ \\
      & 1st \& 2nd-phase  & $74.79$&$72.58$&$70.27$&\cellcolor{gray!30}$62.16$ \\
    \bottomrule
    \end{tabular}%
    \caption{Ablation study of the prototype alignment regularization. mAP($\%$) is adopted as metric. }
    \label{tab:regularization}%
  \end{table}

\section{Conclusion}

  In this work, we present a novel model PML-CD to apply curriculum-based disambiguation on PML in a calibration way. We derive the optimal weighting function from existing curriculum-based PML methods, and provide an insight that assigning curriculum weights for samples is equivalent to the calibration of the model's predicted confidences. Guided by the optimal weighting function, we design a calibrator based on the general relations between confidences histogram and calibration curve, which can provide high quality calibration on unseen dataset after pre-training, therefore serving as an adaptive weighting function on different training condition. We also propose the prototype alignment regularization to improve model's robustness during the training. Comparison experiments have been conducted to verify the effectiveness of PML-CD. 

\section*{Acknowledgements}

  This research work is supported in part by National Natural Science Foundation of China (Grant 62106044 and Grant U24A20322), in part by the Big Data Computing Center of Southeast University.


\bibliography{aaai25}

@article{PML-NI,
  title={Partial multi-label learning with noisy label identification},
  author={Xie, Ming-Kun and Huang, Sheng-Jun},
  journal={IEEE Transactions on Pattern Analysis and Machine Intelligence},
  volume={44},
  number={7},
  pages={3676--3687},
  year={2021},
  publisher={IEEE}
}

@inproceedings{PML-LRS,
  title={Partial multi-label learning by low-rank and sparse decomposition},
  author={Sun, Lijuan and Feng, Songhe and Wang, Tao and Lang, Congyan and Jin, Yi},
  booktitle={Proceedings of the AAAI conference on artificial intelligence},
  volume={33},
  number={01},
  pages={5016--5023},
  year={2019}
}

@inproceedings{PML-LD,
  title={Partial multi-label learning with label distribution},
  author={Xu, Ning and Liu, Yun-Peng and Geng, Xin},
  booktitle={Proceedings of the AAAI conference on artificial intelligence},
  volume={34},
  number={04},
  pages={6510--6517},
  year={2020}
}

@inproceedings{fPML,
  title={Feature-induced partial multi-label learning},
  author={Yu, Guoxian and Chen, Xia and Domeniconi, Carlotta and Wang, Jun and Li, Zhao and Zhang, Zili and Wu, Xindong},
  booktitle={2018 IEEE international conference on data mining (ICDM)},
  pages={1398--1403},
  year={2018},
  organization={IEEE}
}

@article{PLAIN,
  title={Deep partial multi-label learning with graph disambiguation},
  author={Wang, Haobo and Yang, Shisong and Lyu, Gengyu and Liu, Weiwei and Hu, Tianlei and Chen, Ke and Feng, Songhe and Chen, Gang},
  journal={arXiv preprint arXiv:2305.05882},
  year={2023}
}

@article{PML-CDCR,
  title={A Deep Model for Partial Multi-label Image Classification with Curriculum-based Disambiguation},
  author={Sun, Feng and Xie, Ming-Kun and Huang, Sheng-Jun},
  journal={Machine Intelligence Research},
  volume={21},
  number={4},
  pages={801--814},
  year={2024},
  doi={10.1007/s11633-023-1439-3}
}

@article{UNM,
  title={UNM: A Universal Approach for Noisy Multi-label Learning},
  author={Chen, Jia-Yao and Li, Shao-Yuan and Huang, Sheng-Jun and Chen, Songcan and Wang, Lei and Xie, Ming-Kun},
  journal={IEEE Transactions on Knowledge and Data Engineering},
  year={2024},
  publisher={IEEE}
}

@inproceedings{ASL,
  title={Asymmetric loss for multi-label classification},
  author={Ridnik, Tal and Ben-Baruch, Emanuel and Zamir, Nadav and Noy, Asaf and Friedman, Itamar and Protter, Matan and Zelnik-Manor, Lihi},
  booktitle={Proceedings of the IEEE/CVF international conference on computer vision},
  pages={82--91},
  year={2021}
}

@article{Query2Label,
  title={Query2label: A simple transformer way to multi-label classification},
  author={Liu, Shilong and Zhang, Lei and Yang, Xiao and Su, Hang and Zhu, Jun},
  journal={arXiv preprint arXiv:2107.10834},
  year={2021}
}

@inproceedings{ML-Decoder,
  title={Ml-decoder: Scalable and versatile classification head},
  author={Ridnik, Tal and Sharir, Gilad and Ben-Cohen, Avi and Ben-Baruch, Emanuel and Noy, Asaf},
  booktitle={Proceedings of the IEEE/CVF winter conference on applications of computer vision},
  pages={32--41},
  year={2023}
}

@inproceedings{GAELLE,
  title={Self-Supervised Multi-Label Classification with Global Context and Local Attention},
  author={Chen, Chun-Yen and Yeh, Mei-Chen},
  booktitle={Proceedings of the 2024 International Conference on Multimedia Retrieval},
  pages={934--942},
  year={2024}
}

@inproceedings{MentorNet,
  title={Mentornet: Learning data-driven curriculum for very deep neural networks on corrupted labels},
  author={Jiang, Lu and Zhou, Zhengyuan and Leung, Thomas and Li, Li-Jia and Fei-Fei, Li},
  booktitle={International conference on machine learning},
  pages={2304--2313},
  year={2018},
  organization={PMLR}
}

@inproceedings{LL-R,
  title={Large loss matters in weakly supervised multi-label classification},
  author={Kim, Youngwook and Kim, Jae Myung and Akata, Zeynep and Lee, Jungwoo},
  booktitle={Proceedings of the IEEE/CVF Conference on Computer Vision and Pattern Recognition},
  pages={14156--14165},
  year={2022}
}

@inproceedings{SoftMatch,
  title={SoftMatch: Addressing the Quantity-Quality Trade-off in Semi-Supervised Learning},
  author={Chen, Hao and Tao, Ran and Fan, Yue and Wang, Yidong and Wang, Jindong and Schiele, Bernt and Xie, Xing and Raj, Bhiksha and Savvides, Marios},
  booktitle={International Conference on Learning Representations},
  year={2023}
}

@inproceedings{CLC,
  title={CLC: Noisy label correction via curriculum learning},
  author={Lee, Jaeyoon and Lim, Hyuntak and Chung, Ki-Seok},
  booktitle={2021 IEEE Symposium Series on Computational Intelligence (SSCI)},
  pages={1--7},
  year={2021},
  organization={IEEE}
}

@inproceedings{DynaCor,
  title={Learning Discriminative Dynamics with Label Corruption for Noisy Label Detection},
  author={Kim, Suyeon and Lee, Dongha and Kang, SeongKu and Chae, Sukang and Jang, Sanghwan and Yu, Hwanjo},
  booktitle={Proceedings of the IEEE/CVF Conference on Computer Vision and Pattern Recognition},
  pages={22477--22487},
  year={2024}
}

@inproceedings{CenterLoss,
  title={A discriminative feature learning approach for deep face recognition},
  author={Wen, Yandong and Zhang, Kaipeng and Li, Zhifeng and Qiao, Yu},
  booktitle={Computer vision--ECCV 2016: 14th European conference, amsterdam, the netherlands, October 11--14, 2016, proceedings, part VII 14},
  pages={499--515},
  year={2016},
  organization={Springer}
}

@inproceedings{PiCO,
  title={Pico: Contrastive label disambiguation for partial label learning},
  author={Wang, Haobo and Xiao, Ruixuan and Li, Yixuan and Feng, Lei and Niu, Gang and Chen, Gang and Zhao, Junbo},
  booktitle={International conference on learning representations},
  year={2022}
}

@article{PiCO-plus,
  title={Pico+: Contrastive label disambiguation for robust partial label learning},
  author={Wang, Haobo and Xiao, Ruixuan and Li, Yixuan and Feng, Lei and Niu, Gang and Chen, Gang and Zhao, Junbo},
  journal={arXiv preprint arXiv:2201.08984},
  year={2022}
}

@article{EmergingMLL,
  title={The emerging trends of multi-label learning},
  author={Liu, Weiwei and Wang, Haobo and Shen, Xiaobo and Tsang, Ivor W},
  journal={IEEE transactions on pattern analysis and machine intelligence},
  volume={44},
  number={11},
  pages={7955--7974},
  year={2021},
  publisher={IEEE}
}

@article{EarlyStopping,
  title={Understanding and improving early stopping for learning with noisy labels},
  author={Bai, Yingbin and Yang, Erkun and Han, Bo and Yang, Yanhua and Li, Jiatong and Mao, Yinian and Niu, Gang and Liu, Tongliang},
  journal={Advances in Neural Information Processing Systems},
  volume={34},
  pages={24392--24403},
  year={2021}
}

@inproceedings{a-closer-look,
  title={A closer look at memorization in deep networks},
  author={Arpit, Devansh and Jastrz{\k{e}}bski, Stanis{\l}aw and Ballas, Nicolas and Krueger, David and Bengio, Emmanuel and Kanwal, Maxinder S and Maharaj, Tegan and Fischer, Asja and Courville, Aaron and Bengio, Yoshua and others},
  booktitle={International conference on machine learning},
  pages={233--242},
  year={2017},
  organization={PMLR}
}

@inproceedings{RoCL,
  title={Robust curriculum learning: from clean label detection to noisy label self-correction},
  author={Zhou, Tianyi and Wang, Shengjie and Bilmes, Jeff},
  booktitle={International Conference on Learning Representations},
  year={2020}
}

@inproceedings{CTRR,
  title={On learning contrastive representations for learning with noisy labels},
  author={Yi, Li and Liu, Sheng and She, Qi and McLeod, A Ian and Wang, Boyu},
  booktitle={Proceedings of the IEEE/CVF conference on computer vision and pattern recognition},
  pages={16682--16691},
  year={2022}
}

@inproceedings{cnn-rnn,
  title={Cnn-rnn: A unified framework for multi-label image classification},
  author={Wang, Jiang and Yang, Yi and Mao, Junhua and Huang, Zhiheng and Huang, Chang and Xu, Wei},
  booktitle={Proceedings of the IEEE conference on computer vision and pattern recognition},
  pages={2285--2294},
  year={2016}
}

@inproceedings{Dash,
  title={Dash: Semi-supervised learning with dynamic thresholding},
  author={Xu, Yi and Shang, Lei and Ye, Jinxing and Qian, Qi and Li, Yu-Feng and Sun, Baigui and Li, Hao and Jin, Rong},
  booktitle={International conference on machine learning},
  pages={11525--11536},
  year={2021},
  organization={PMLR}
}

@inproceedings{DRAMA,
  title={Discriminative and Correlative Partial Multi-Label Learning.},
  author={Wang, Haobo and Liu, Weiwei and Zhao, Yang and Zhang, Chen and Hu, Tianlei and Chen, Gang},
  booktitle={IJCAI},
  pages={3691--3697},
  year={2019}
}

@inproceedings{Easy-samples-first,
  title={Easy samples first: Self-paced reranking for zero-example multimedia search},
  author={Jiang, Lu and Meng, Deyu and Mitamura, Teruko and Hauptmann, Alexander G},
  booktitle={Proceedings of the 22nd ACM international conference on Multimedia},
  pages={547--556},
  year={2014}
}

@inproceedings{C-Tran,
  title={General multi-label image classification with transformers},
  author={Lanchantin, Jack and Wang, Tianlu and Ordonez, Vicente and Qi, Yanjun},
  booktitle={Proceedings of the IEEE/CVF conference on computer vision and pattern recognition},
  pages={16478--16488},
  year={2021}
}

@article{GLADE,
  title={Multi-view partial multi-label learning via graph-fusion-based label enhancement},
  author={Xu, Ning and Wu, Yong-Di and Qiao, Congyu and Ren, Yi and Zhang, Minxue and Geng, Xin},
  journal={IEEE Transactions on Knowledge and Data Engineering},
  volume={35},
  number={11},
  pages={11656--11667},
  year={2022},
  publisher={IEEE}
}

@article{GLC,
  title={Global-local label correlation for partial multi-label learning},
  author={Sun, Lijuan and Feng, Songhe and Liu, Jun and Lyu, Gengyu and Lang, Congyan},
  journal={IEEE Transactions on Multimedia},
  volume={24},
  pages={581--593},
  year={2021},
  publisher={IEEE}
}

@article{PAKS,
  title={Prior knowledge regularized self-representation model for partial multilabel learning},
  author={Lyu, Gengyu and Feng, Songhe and Jin, Yi and Wang, Tao and Lang, Congyan and Li, Yidong},
  journal={IEEE Transactions on Cybernetics},
  volume={53},
  number={3},
  pages={1618--1628},
  year={2021},
  publisher={IEEE}
}

@article{PML-SALC,
  title={Partial multi-label learning based on sparse asymmetric label correlations},
  author={Zhao, Peng and Zhao, Shiyi and Zhao, Xuyang and Liu, Huiting and Ji, Xia},
  journal={Knowledge-Based Systems},
  volume={245},
  pages={108601},
  year={2022},
  publisher={Elsevier}
}

@inproceedings{Scene-aware,
  title={Scene-aware label graph learning for multi-label image classification},
  author={Zhu, Xuelin and Liu, Jian and Liu, Weijia and Ge, Jiawei and Liu, Bo and Cao, Jiuxin},
  booktitle={Proceedings of the IEEE/CVF International Conference on Computer Vision},
  pages={1473--1482},
  year={2023}
}

@article{kumar2010self,
  title={Self-paced learning for latent variable models},
  author={Kumar, M and Packer, Benjamin and Koller, Daphne},
  journal={Advances in neural information processing systems},
  volume={23},
  year={2010}
}

@article{Survey-curriculum-learning,
  title={A survey on curriculum learning},
  author={Wang, Xin and Chen, Yudong and Zhu, Wenwu},
  journal={IEEE transactions on pattern analysis and machine intelligence},
  volume={44},
  number={9},
  pages={4555--4576},
  year={2021},
  publisher={IEEE}
}

@inproceedings{Curriculum-learning-2009,
  title={Curriculum learning},
  author={Bengio, Yoshua and Louradour, J{\'e}r{\^o}me and Collobert, Ronan and Weston, Jason},
  booktitle={Proceedings of the 26th annual international conference on machine learning},
  pages={41--48},
  year={2009}
}

@inproceedings{ML-GCN,
  title={Multi-label image recognition with graph convolutional networks},
  author={Chen, Zhao-Min and Wei, Xiu-Shen and Wang, Peng and Guo, Yanwen},
  booktitle={Proceedings of the IEEE/CVF conference on computer vision and pattern recognition},
  pages={5177--5186},
  year={2019}
}

@article{ML-AGCN,
  title={Multi-label image classification using adaptive graph convolutional networks: from a single domain to multiple domains},
  author={Singh, Inder Pal and Ghorbel, Enjie and Oyedotun, Oyebade and Aouada, Djamila},
  journal={Computer Vision and Image Understanding},
  volume={247},
  pages={104062},
  year={2024},
  publisher={Elsevier}
}

@article{VOC,
  title={The pascal visual object classes (voc) challenge},
  author={Everingham, Mark and Van Gool, Luc and Williams, Christopher KI and Winn, John and Zisserman, Andrew},
  journal={International journal of computer vision},
  volume={88},
  pages={303--338},
  year={2010},
  publisher={Springer}
}

@inproceedings{COCO,
  title={Microsoft coco: Common objects in context},
  author={Lin, Tsung-Yi and Maire, Michael and Belongie, Serge and Hays, James and Perona, Pietro and Ramanan, Deva and Doll{\'a}r, Piotr and Zitnick, C Lawrence},
  booktitle={Computer Vision--ECCV 2014: 13th European Conference, Zurich, Switzerland, September 6-12, 2014, Proceedings, Part V 13},
  pages={740--755},
  year={2014},
  organization={Springer}
}

@inproceedings{NUS-WIDE,
  title={Nus-wide: a real-world web image database from national university of singapore},
  author={Chua, Tat-Seng and Tang, Jinhui and Hong, Richang and Li, Haojie and Luo, Zhiping and Zheng, Yantao},
  booktitle={Proceedings of the ACM international conference on image and video retrieval},
  pages={1--9},
  year={2009}
}

@article{CUB,
  title={The caltech-ucsd birds-200-2011 dataset},
  author={Wah, Catherine and Branson, Steve and Welinder, Peter and Perona, Pietro and Belongie, Serge},
  year={2011},
  publisher={California Institute of Technology}
}

@inproceedings{ResNet,
  title={Deep residual learning for image recognition},
  author={He, Kaiming and Zhang, Xiangyu and Ren, Shaoqing and Sun, Jian},
  booktitle={Proceedings of the IEEE conference on computer vision and pattern recognition},
  pages={770--778},
  year={2016}
}

@inproceedings{Randaugment,
  title={Randaugment: Practical automated data augmentation with a reduced search space},
  author={Cubuk, Ekin D and Zoph, Barret and Shlens, Jonathon and Le, Quoc V},
  booktitle={Proceedings of the IEEE/CVF conference on computer vision and pattern recognition workshops},
  pages={702--703},
  year={2020}
}

@inproceedings{NLR,
  title={Noisy Label Removal for Partial Multi-Label Learning},
  author={Yang, Fuchao and Jia, Yuheng and Liu, Hui and Dong, Yongqiang and Hou, Junhui},
  booktitle={Proceedings of the 30th ACM SIGKDD Conference on Knowledge Discovery and Data Mining},
  pages={3724--3735},
  year={2024}
}

@inproceedings{CEL,
  title={Mixed Blessing: Class-Wise Embedding guided Instance-Dependent Partial Label Learning},
  author={Yang, Fuchao and Cheng, Jianhong and Liu, Hui and Dong, Yongqiang and Jia, Yuheng and Hou, Junhui},
  booktitle={Proceedings of the 31st ACM SIGKDD Conference on Knowledge Discovery and Data Mining},
  pages={1763--1772},
  year={2025},
  doi={10.1145/3690624.3709276}
}

\end{document}